\documentclass[letterpaper, 10 pt, conference]{ieeeconf}  

\IEEEoverridecommandlockouts                              

\usepackage{balance}
\usepackage{amsmath,amsfonts,amssymb}
\usepackage{mathtools}
\usepackage{commath}
\usepackage{bm}
\usepackage{color}
\usepackage{multirow}
\usepackage{tabularx}
\usepackage[flushleft]{threeparttable}
\usepackage{booktabs}
\usepackage{url}
\usepackage{algorithm}
\usepackage{algorithmic}
\usepackage{cite}
\usepackage{makecell}
\usepackage[switch]{lineno}

\title{\LARGE \bf
Poses as Queries: Image-to-LiDAR Map Localization with Transformers
}

\author{Jinyu Miao$^{1}$, Kun Jiang$^{1,*}$, Yunlong Wang$^{1}$, Tuopu Wen$^{1}$, Zhongyang Xiao$^{2}$, Zheng Fu$^{1}$, \\ Mengmeng Yang$^{1,*}$, Maolin Liu$^{1}$, Diange Yang$^{1,*}$ 
\thanks{This work was supported in part by the National Natural Science Foundation of China under Grants U22A20104, and Beijing Municipal Science and Technology Commission (Grant No.Z221100008122011).}
\thanks{$^{1}$Jinyu Miao, Kun Jiang, Yunlong Wang, Tuopu Wen, Zheng Fu, Mengmeng Yang, Maolin Liu, and Diange Yang are with the School of Vehicle and Mobility, Tsinghua University, Beijing, 100084, China. 
        {\tt\small jinyu.miao97@gmail.com}}%
\thanks{$^{2}$Zhongyang Xiao is with Autonomous Driving Division of NIO Inc., Beijing, China}%
\thanks{$^{*}$Corresponding author: Diange Yang, Kun Jiang and Mengmeng Yang}
}

\begin{document}

\maketitle
\thispagestyle{empty}
\pagestyle{empty}

\begin{abstract}
High-precision vehicle localization with commercial setups is a crucial technique for high-level autonomous driving tasks. Localization with a monocular camera in LiDAR map is a newly emerged approach that achieves promising balance between cost and accuracy, but estimating pose by finding correspondences between such cross-modal sensor data is challenging, thereby damaging the localization accuracy. In this paper, we address the problem by proposing a novel Transformer-based neural network to register 2D images into 3D LiDAR map in an end-to-end manner. 
Poses are implicitly represented as high-dimensional feature vectors called \textit{pose queries} and can be iteratively updated by interacting with the retrieved relevant information from cross-model features using attention mechanism in a proposed POse Estimator Transformer (POET) module.
Moreover, we apply a multiple hypotheses aggregation method that estimates the final poses by performing parallel optimization on multiple randomly initialized \textit{pose queries} to reduce the network uncertainty.
Comprehensive analysis and experimental results on public benchmark conclude that the proposed image-to-LiDAR map localization network could achieve state-of-the-art performances in challenging cross-modal localization tasks.  
\end{abstract}

\section{Introduction}
High-precision vehicle localization services as a prerequisite in high-level autonomous driving system for its ability to provide real-time poses in a pre-built map. The given poses can be applied to load environmental information from map, which boost the performance of subsequent navigation, decision making, and control for autonomous vehicles. 

Traditional map-based localization algorithm can be roughly categorized into two classes based on the utilized sensors, namely, visual localization and LiDAR localization. Such localization algorithms are commonly constructed as two-stage hierarchical frameworks, that is, place recognition and metric pose estimation \cite{hfnet,bvmatch}. 
The place recognition stage firstly retrieves geographically neighboring keyframes by visual descriptor \cite{dbow-tro, netvlad} or LiDAR descriptor \cite{scancontext, scancontext++}. 
Then the metric pose estimation stage performs map matching to recover precise pose. 
Visual localization generally matches feature descriptors between current frame and visual landmarks in the map, and then solves perspective-n-points (PnP) problem in a random sample consensus (RANSAC) \cite{ransac} circulation by minimizing re-projection errors. Visual methods only need low-cost camera, but its performance heavily relies on the accuracy of feature matching and the quality of visual map. 
LiDAR localization aligns the geometry or distribution between current scan and point clouds in the map by using iterative closest point (ICP) series \cite{icp,p2licp,gicp} or normal distribution transform (NDT) algorithm \cite{ndt,3dndt}. As a comparison, LiDAR map provides more dense and accurate representation of scene, but the alignment is more challenging and requires geometrical good initial value. And high-precision LiDAR sensor is costly and needs high power consumption. 
For the sake of economical vehicles, localization algorithms with low-cost sensor suite are needed to be developed \cite{xiao_itsc,xiao_sensors,wen_iv,wen_tits}. 
As a newly emerged method, visual localization in LiDAR map only need monocular camera in the localization stage while they could sufficiently utilize accurate LiDAR map, which seems a potential excellent attempt about the balance between localization accuracy and sensor consumption \cite{vloc_vo,vloc_line,vloc_stereo}. 
However, the inherent difference of the modalities challenges matching between cross-modal data in the localization algorithm, which can be probably solved by camera-LiDAR calibration methods.

Target-less extrinsic calibration between monocular camera and LiDAR has been well studied for a long time. Recently, some proposals apply deep learning method to directly regress the transformation between camera and LiDAR \cite{regnet, deepi2p, efghnet, lccnet}. These methods convert point clouds from LiDAR scan into depth images, and apply convolutional neural network (CNN) to extract features from both sensor data so as to regress the rigid transformation between two sensors. 
Following a similar way, Cattaneo \textit{et al.} presented CMRNet \cite{cmrnet} for visual localization in LiDAR map that the only technical difference is that the depth image fed into CMRNet \cite{cmrnet} is generated from LiDAR map not a single LiDAR scan. Later, HyperMap \cite{hypermap} tends to save the map storage. 
However, all these methods simply build pose estimators by some stacked convolutional layers and fully connection layers, and directly regress pose in one shot. Such simple and unreasonable networks cannot fully exploit matching information and result in unpleasant localization performance.

In this paper, we also follow the camera-to-LiDAR map localization method to achieve low-cost and high-precision vehicle localization. 
To improve the localization accuracy, we present a Transformer-based neural network and propose to implicitly represent poses as high-dimensional feature vectors, named as \textit{pose queries} in this work. Especially, we design a novel \textbf{PO}se \textbf{E}stimator \textbf{T}ransformer (POET) module where the \textit{pose queries} can be iteratively updated by retrieving relevant matching information from the cost volume between cross-modal features. Benefited by the proposed POET module, our network could achieve a significantly improved localization accuracy when integrated in a image-to-LiDAR map localization pipeline. The primary contributions are summarized as:

\begin{itemize}
    \item A novel POET module is proposed where poses are implicitly represented as high-dimensional feature vectors and can be updated as queries in Transformer. By applying the module, precise pose estimation with monocular camera in LiDAR map can be achieved.
    \item A multiple hypotheses aggregation method is applied to reduce the uncertainty of the proposed networks. We perform parallel optimization on several randomly initialized \textit{pose queries}, and aggregate the optimized \textit{pose queries} to estimate more stably.
    \item The proposed network with POET modules is integrated into an iterative image-to-LiDAR map localization system. Experimental results show our method could achieve high localization accuracy.
\end{itemize}

The remainder of the paper is organized as follows. We review relevant works in Section \ref{review}. The proposed network with the POET module and its training scheme are introduced in details in the section \ref{method}. Comprehensive experiments to demonstrate the effectiveness of the proposed network are provided in the section \ref{method}. Finally, section \ref{conclusion} ends the paper with conclusion.

\section{Related Works}
\label{review}

As a dispensable component in autonomous systems, localization algorithms have been developed for many decades. We only introduce the works most relevant to this paper, namely, visual localization and camera-to-LiDAR calibration.

\subsection{Visual only Localization} 
The lost cost of monocular camera makes visual localization a popular stride to be developed for both academic and industrial societies. Most of them generally follow a coarse-to-fine scheme, \textit{i.e.}, hierarchical localization \cite{hfnet}. The coarse stage extracts image global features and then performs place recognition to retrieve historical images \cite{dbow-tro,netvlad}. But the retrieved images only provide rough pose approximation for localization. Thus, A fine stage to recover precise pose need to conducted. 
In \cite{models}, authors claimed that 3D models are not strictly necessary for visual localization and they refined the poses by a weighted combination of the poses of retrieved images. 
However, since the poses cannot ensure to be linear to the features and maintaining numerous historical images also costs an unwieldy amount of memory, localization by matching between 2D images and 3D scene model is still the most popular choice. 
HLoc \cite{hfnet} matches sparse local features whereas InLoc \cite{inloc} performs dense CNN matching. Then the 2D-2D feature matches are converted to the 2D-3D correspondences between 2D pixels and 3D visual landmarks in the map so that they can estimate a precise pose with P3P-RANSAC algorithm \cite{p3p,ransac}. 
Some other works perform map matching between recognized semantic-level elements and vectorized high-definition map (HD Map) and achieve commercial localization with only monocular camera \cite{wen_iv,wen_tits,xiao_itsc,xiao_sensors}. 
Recently, some deep learning-based absolute pose regression algorithms have been developed \textit{e.g.}, PoseNet \cite{posenet} and CaTiLoc \cite{catiloc}, but they are hard to hold a high localization accuracy in large-scale scene. 
Among these visual only localization methods, the scene map can be 3D models with visual descriptors \cite{hfnet,inloc}, geo-tagged keyframes \cite{dbow-tro,netvlad} or neural network \cite{posenet,catiloc}.

\subsection{Visual localization in LiDAR map}
Generally speaking, visual map is hard to achieve a comparable accuracy to the LiDAR map, thus HD Maps usually contains the point clouds scanned by 3D LiDAR sensors. Also, raw point cloud map does not have the necessity to save features, reducing the storage requirements. Some visual localization algorithms with LiDAR map have been developed in last decade under this context.
Since matching features between such cross-modal sensor data is challenging, some solutions turn to match geometry. 
Caselitz \textit{et al.} proposed to reconstruct a 3D local map by a visual odometry (VO) so that the local map can be aligned to global LiDAR map \cite{vloc_vo}. Later, Yu \textit{et al.} extracted geometric lines and involved 2D-3D line correspondences into iteration optimization \cite{vloc_line}. Kim \textit{et al.} exploited a stereo camera to obtain depth of current viewpoint and then match it with map \cite{vloc_stereo}. All of these methods have a requisite to run a VO thread to lift 2D points to 3D points or give a initial pose prediction, which is quite computational consuming and limits the application to a continual execution manner. 
Therefore, some deep learning-based methods to find the correspondences between cross-modal data are developed. CMRNet \cite{cmrnet} obtains the cost volume between the image feature and LiDAR feature by a optical flow network \cite{pwcnet}, and then regresses the pose of monocular camera with regard to the LiDAR map. Then, the same author proposed CMRNet++ \cite{cmrnet++} to predict correspondences between image and LiDAR so that the cross-modal localization could be solved by a EPnP-RANSAC way \cite{epnp,ransac}. Chang \textit{et al.} compressed the LiDAR map to reduce map size by 87-94$\%$ while achieving comparable or better accuracy. 
We also follow this way to estimate poses in an end-to-end manner and make efforts to improve the localization accuracy to centimeter-level.

\begin{figure*}[!t]
    \centering
    \includegraphics[width=0.97\linewidth]{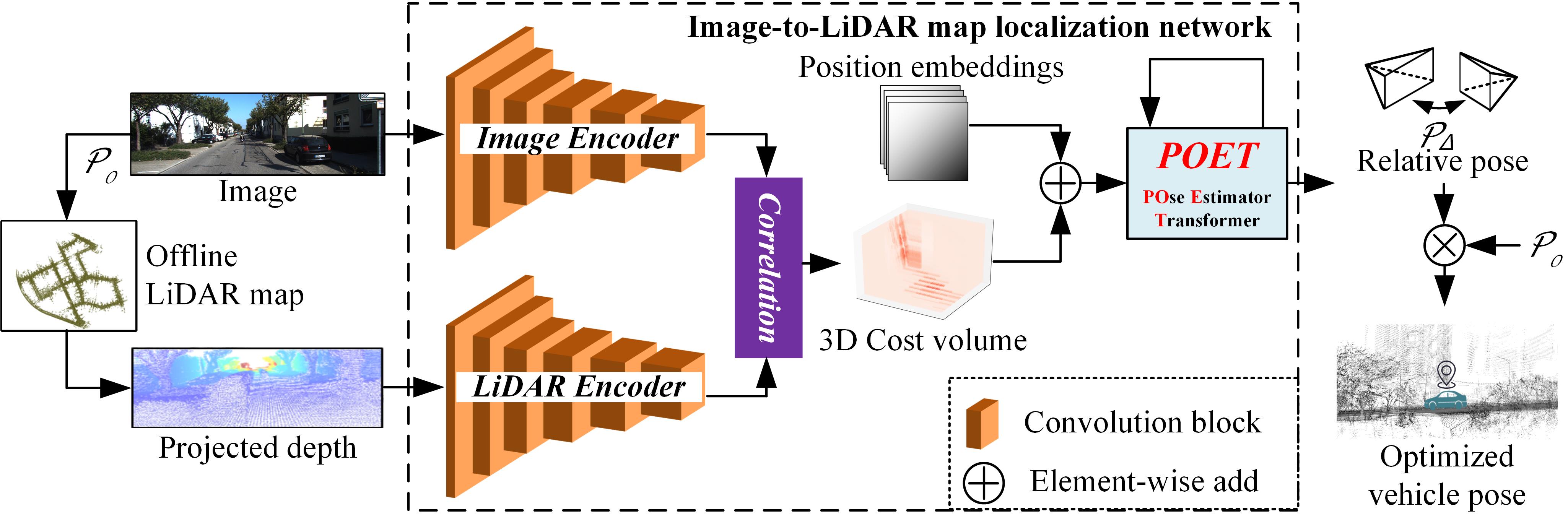}
    \caption{The overall structure of the proposed image-to-LiDAR map localization network.}
    \label{fig:framework}
\end{figure*}

\subsection{Camera-to-LiDAR Calibration}
Visual localization in LiDAR map is technically similar to target-less camera-to-LiDAR extrinsic calibration, the only difference is that the localization methods need a clip of LiDAR point cloud map. 
Early proposed calibration method aligns strategies, called hand-eye extrinsic calibration, to estimate the rigid transformation from camera to LiDAR \cite{handeye}. Such a strategy needs the vehicle to run in a $\infty$ shaped trajectory and cannot operate in online manner. 
With the development of detection and segmentation methods, some methods estimate the extrinsic parameters by minimizing re-projection errors between 2D extracted poles/signs and 3D vectorized map elements \cite{opencalib}. These methods cannot adopt to visual localization task due to the sparse map elements in real scenarios. 
Recently, deep learning-based methods have raised. As the first work in this line, RegNet \cite{regnet} concatenates image features and LiDAR features and then regresses the calibration results via the fused features. Different strategy is proposed in DeepI2P \cite{deepi2p} that it designs a classification network to label whether the projection of each 3D point is within or beyond the camera frustum, then these labeled points are used to estimate extrinsic parameters by an inverse camera projection solver. Jeon \textit{et al.} proposed EFGHNet \cite{efghnet} to estimate the transformation in a divide-and-conquer strategy with a two-phase structure, thereby leading to better accuracy. LCCNet \cite{lccnet} applies a similar network to CMRNet \cite{cmrnet} that it utilizes two parallel branch to separately extract high-dimensional features from RGB image and depth image and calculate a 3D cost volume by a correlation layer proposed in PWC-Net \cite{pwcnet}, then the extrinsic parameters are regressed. The idea behind these works that convert cross-modal data to CNN features and estimate pose using neural networks motivates our innovations.

\section{Methodology}
\label{method}

In this section, we will describe the structure and training scheme of the proposed image-to-LiDAR map localization network in details. 

\subsection{Overall Structure}
As shown in the Fig.\ref{fig:framework}, the proposed network is fed by a RGB image $\mathcal{I} \in \mathbb{R}^{H \times W \times 3}$ and a projected depth image $\mathcal{L} \in \mathbb{R}^{H \times W}$. The projected depth image is generated by re-projecting the neighboring point clouds in the LiDAR map onto a virtual image plane on a given initial pose $\mathcal{P}_0$. Then, the image $\mathcal{I}$ and the depth $\mathcal{L}$ are processed by corresponding encoder to get high-dimensional features respectively. Applying a correlation module, we get a cost volume between image and LiDAR features. We then add positional embedding to the cost volume and feed the flatted cost volume into the proposed POET module. And the relative pose $\mathcal{P}_\triangle$ between the viewpoint of the image $\mathcal{I}$ and the initial pose $\mathcal{P}_0$ can be estimated.

\subsection{Encoder and Correlation Modules}
We follow CMRNet \cite{cmrnet} and LCCNet \cite{lccnet} to construct the encoder and correlation modules. These modules aim to efficiently extract robust features and get matching information, so called cost volume, between cross-modal sensor data. 

The image encoder is composed by six convolutional blocks that lifts a RGB image with resolution $(H,W)$ into a feature map $\mathcal{F}_\mathcal{I} \in \mathbb{R}^{H_c \times W_c \times D_c}, H_c=H/64, W_c=W/64, D_c=196$. 
Each convolutional block in the encoder has similar structure including three alternatively arranged convolutional layers and non-linear activation layers. All the convolutional layer apply $3 \times 3$ convolutional kernels. The first convolutional layer in each block sets stride to be 2 and others are 1 so that feature map went through each block will be compressed by half as original ones. The output channels of convolutional kernels in each convolutional block are respectively 16, 32, 64, 96, 128, and 196. 
The non-linear activation used in this work is leaky rectified linear unit (LeakyReLU) \cite{leakyrelu} with slope 0.1 for negative values. 
The LiDAR encoder has almost the same structure as image encoder and the only difference is that the first convolutional layer is fed by an one-channel depth image. 
By applying image and LiDAR encoder, we can get two feature maps $\mathcal{F}_\mathcal{I}, \mathcal{F}_\mathcal{L}$ respectively.

Directly calculate the matching cost between each feature vector in $\mathcal{F}_\mathcal{I}$ and $\mathcal{F}_\mathcal{L}$ will results in unbearable computation burden, and the resulting 4D cost volume is also hard to be processed. 
Therefore, we apply a more efficient way used in PWC-Net \cite{pwcnet} to calculate 3D cost volume. The matching cost is defined as the the correlation between image features and LiDAR features:

\begin{equation}
\label{equ:1}
    c(x_I, y_L)=\frac{1}{N}{(\mathcal{F}_\mathcal{I}(x_I))}^T(\mathcal{F}_\mathcal{L}(y_L))
\end{equation}
where $N$ is the length of the feature vector $\mathcal{F}_\mathcal{*}(x_*)$, $x_I,y_L$ are the indexes of features in the $\mathcal{F}_\mathcal{I}$ and $\mathcal{F}_\mathcal{L}$ respectively. 
Since the initial pose $\mathcal{P}_0$ is assumed to be near the ground truth vehicle pose, the displacement between two feature maps will be limited. Calculating costs between a feature in $\mathcal{F}_\mathcal{I}$ and all the features in $\mathcal{F}_\mathcal{L}$ is unnecessary. 
Thus, we compute a partial cost volume within $d$ pixels, \textit{i.e.}, ${|x-y|}_\infty \le d$, which corresponds to a maximum displacement as many as $d \cdot 2^{6}$ pixels at full resolution of original images. 
In this work, we set $d$ to 4, so the dimension of the resulting 3D cost volume $\mathcal{C}_{\mathcal{I},\mathcal{L}}$ is $D_{cv}={(d+1)}^2 \times H_c \times W_c$. 
Such a cost volume can be seen as the matching information between the image $\mathcal{I}$ and the map, which is similar to traditional PnP-based localization methods but we take use of more comprehensive information.

\subsection{POse Estimator Transformer (POET)}

{
\setlength{\parindent}{0cm}
\textbf{Preliminaries: Transformer attention \cite{attention}}
We apply Transformer here to achieve POET module and for better readability, we briefly review Transformer here as background. As the key component in Transformer, attention layers take d-dimensional query vector Q, key vector K, and value vector V as input. The calculation process in an attention layer can be formatted as:
}
\begin{equation}
\label{equ:2}
    \texttt{Attention}(Q,K,V)=\texttt{softmax}(\frac{QK^T}{\sqrt{d}})V
\end{equation}
Intuitively, the query vector Q retrieves related information from the value vector V based on the similarity weight between the query vector Q and the key vector K. 

\begin{figure}[t]
    \centering
    \includegraphics[width=0.97\linewidth]{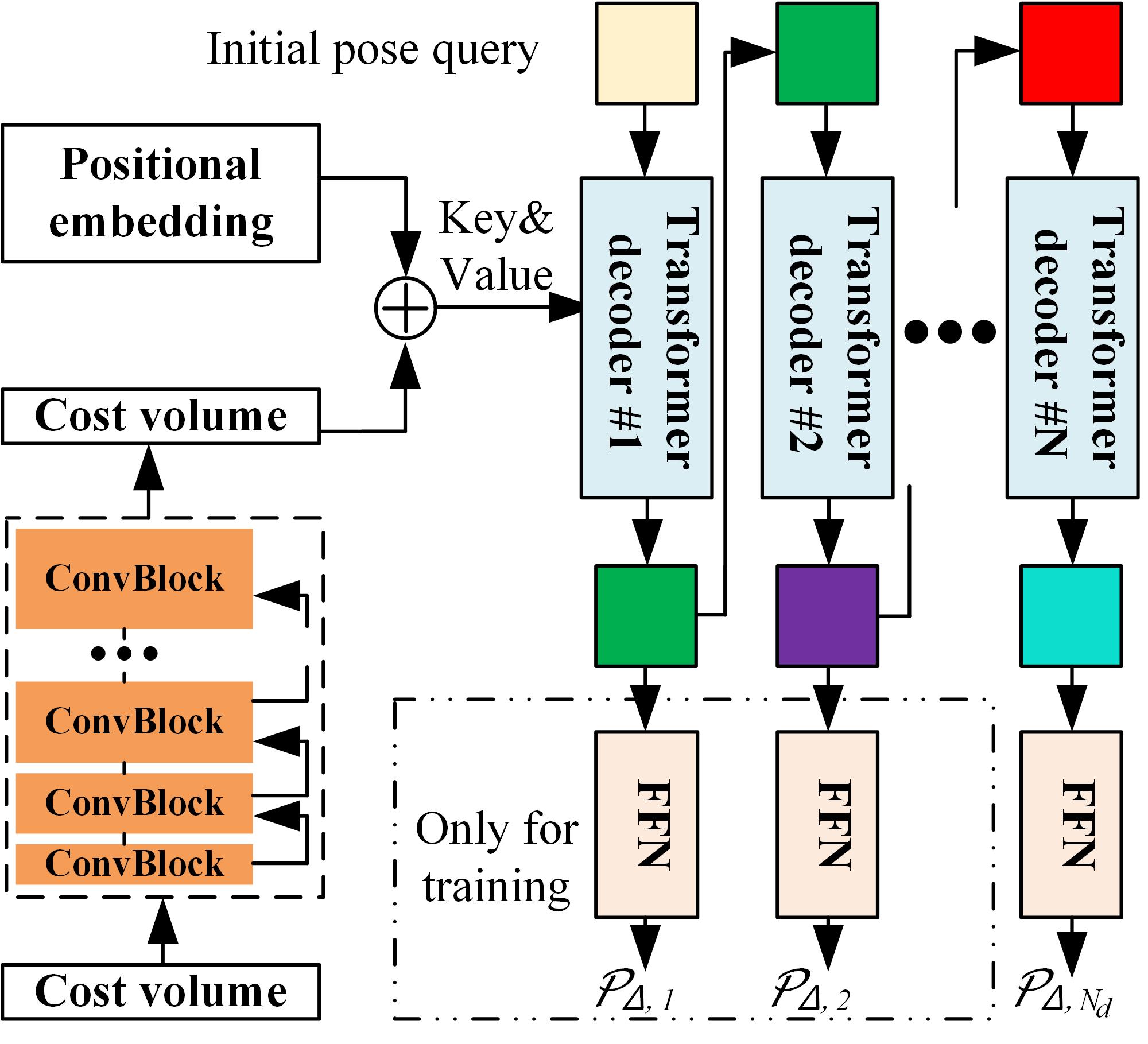}
    \caption{The detailed structure of the proposed pose estimator Transformer (POET) module.}
    \label{fig:poet}
\end{figure}

Regarding the pose estimation module based on the matching information between the image and the LiDAR map, we propose a novel Transformer-based pose estimator POET instead of the vanilla regressor stacked by several convolutional layers and fully connection layers \cite{cmrnet, lccnet, hypermap}. As shown in the Fig. \ref{fig:poet}, POET takes cost volume as input and initializes \textit{pose query}. After iterative updates by related information from the cost volume, the \textit{pose query} is refined to high-precision relative pose between the image $\mathcal{I}$ and initial pose $\mathcal{P}_0$.

Formally, given the cost volume $\mathcal{C}_{\mathcal{I},\mathcal{L}}$, we firstly lift its dimension to $D'_{cv}=256$ by some densely connected convolutional layers \cite{densenet} and then add 2D extension of absolute sinusoidal positional embedding to the cost volume to preserve the position information following \cite{detr}. The positional embedding for $i^{th}$ channel of the cost volume on $(x,y)$ is as follows:

\begin{equation}
\label{equ:3}
    {PE}^i_{x,y}:= \left\{  
             \begin{array}{rcl}  
                 \texttt{sin}(\omega_k \cdot x)&,& i=4k \\  
                 \texttt{cos}(\omega_k \cdot x)&,& i=4k+1\\  
                 \texttt{sin}(\omega_k \cdot y)&,& i=4k+2\\
                 \texttt{cos}(\omega_k \cdot y)&,& i=4k+3
             \end{array}  
            \right.
\end{equation}
where $\omega_k=\frac{1}{10000^{2k/D'_{cv}}}$. Then, the processed cost volume $\mathcal{C}_{\mathcal{I},\mathcal{L}} \in \mathbb{R}^{H_c \times W_c \times D'_{cv}}$ is reorganized to vector format $\overline{\mathcal{C}}=\{\overline{\mathcal{C}}_i\}^{H_c \times W_c}_{i=1}$ where $\overline{\mathcal{C}}_i \in \mathbb{R}^{D'_{cv}}$, which can be seen as $H_c \times W_c$ $D'_{cv}$-dimensional feature vectors. 

In this work, we regard poses as high-dimensional feature vectors and hope they can be updated by related information from the cost volume. Therefore, we randomly initialize a feature vector $\overline{\mathcal{Q}}^0_p \in \mathbb{R}^{D'_{cv}}$ as the implicit representation of the pose, denoted as \textit{pose query}. And we apply DETR \cite{detr} decoder here to update \textit{pose query}. The decoder is composed by alternatively stacked self-attention and cross-attention layer. Self-attention is calculated within the \textit{pose query} $\overline{\mathcal{Q}}_p$ while cross-attention is calculated between the \textit{pose query} $\overline{\mathcal{Q}}_p$ and the processed cost volume $\overline{\mathcal{C}}$. We utilize $N_d$ decoder layers in POET to gradually update the \textit{pose query} and in order to boost the performance of refinement based on prior knowledge, the fed \textit{pose query} of latter decoder layer is the updated \textit{pose query} from former decoder layer as shown in the Fig. \ref{fig:poet}:

\begin{equation}
\label{equ:4}
    \overline{\mathcal{Q}}^{k}_p = \texttt{decoder}_k(\overline{\mathcal{Q}}^{k-1}_p,\overline{\mathcal{C}}), k\in [1,N_d]
\end{equation}

After getting the updated implicit representation of the pose $\overline{\mathcal{Q}}^{*}_p$, each transformer decoder is assigned to a head with two fully connection layers to decode $\overline{\mathcal{Q}}^{*}_p$ to relative pose formatted as 7D vector $\mathcal{P}_{\triangle,*}$, composed by a 3D translation vector $\mathbf{t}$ and a 4D rotation quaternion $\mathbf{q}=[qw,qx,qy,qz]$. 

For efficiency, we also adopt multi-head attention in the transformer attention layer. And the first $N_d-1$ heads are only used for training, we discard them after training is done and only maintain prediction from the last decoder $\mathcal{P}_{\triangle,N_d}$ as the final result of the network during inference.

\subsection{Iterative Pose refinement}

Looking back to the generation of projected depth image $\mathcal{L}$, given an initial camera pose $\mathcal{P}_0$, the point clouds $P_w={[X_w,Y_w,Z_w]}^T$ in the global frame can be transformed into a virtual viewpoint located in $\mathcal{P}_0$:
\begin{equation}
\label{equ:5}
    P_v = {[X_v,Y_v,Z_v,1]}^T = \mathcal{H}(\mathcal{P}_0) \left[\begin{array}{c}
                                                P_w \\
                                                1 
                                              \end{array} \right]
\end{equation}
where $\mathcal{H}(\cdot)$ converts a 7D pose vector to the homogeneous transform matrix in $\texttt{SE}(3)$. Then, according to the known intrinsic $K$ of camera model, projected depth image $\mathcal{L}_{\mathcal{P}_0}$ in viewpoint $\mathcal{P}_0$ can be obtained:
\begin{equation}
\label{equ:6}
    \mathcal{L}_{\mathcal{P}_0}(\pi (P_v, K)) = Z_v
\end{equation}
where $\pi()$ returns 2D projection of 3D points.

After running the proposed network once, we can get an estimated relative pose between the viewpoint of $\mathcal{I}$ and the virtual viewpoint $\mathcal{P}_0$, so a more precise absolute pose can be calculated:
\begin{equation}
\label{equ:7}
    \mathcal{P}_1 = \mathcal{H}^{-1}(\mathcal{H}(\mathcal{P}_{\triangle,N_d})\mathcal{H}(\mathcal{P}_0))
\end{equation}

Using the updated pose $\mathcal{P}_1$ as a new initial pose, a new projected depth image $\mathcal{L}_1$ can be obtained and should be aligned to the image $\mathcal{I}$ with less displacements, which will further boost next estimation step of the proposed network. In this work, we run aforementioned iterative pose refinement three times at most as \cite{cmrnet,hypermap}.

\subsection{Multiple Hypotheses Aggregation for Pose Queries}

We randomly initialize the initial \textit{pose query} in this work that may brings uncertainty in the inference. In the experiments, we found that multiple runs will generate various results and some results are bad, which probably impute to bad initial value. To prevent from such a phenomenon and make the network more stable, we apply a simple but effective way that we utilize multiple \textit{pose queries} in the POET. Formally, the input \textit{pose query} in each Transformer decoder is extended as $\{{}^i\overline{\mathcal{Q}}^*_p\}^{N_q}_{i=1}$ and thus we can get multiple predictions from each POET. We simply average the multiple predictions and fed it into the original head to estimate relative pose. Averaging over multiple \textit{pose queries} will weaken the bad influence caused by some bad hypotheses of \textit{pose query} initialization and thus enhance the stability of the proposed network.

Notice that we only train the network with $N_q=1$ \textit{pose query} and predict with more \textit{pose queries}. Therefore, the applied multiple hypotheses aggregation scheme does not need re-train.

\subsection{Training Scheme}

The training scheme used in this work is similar to CMRNet \cite{cmrnet}. According to the predicted pose of LiDAR SLAM and extrinsic parameters, we can get the aligned point clouds with regard to each image. We then transform the point clouds by an uniformly distributed transformation $\hat{\mathcal{P}_\triangle}$, which can be seen as the localization error of initial pose $\mathcal{P}_0$. We also perform some data augmentation method such as randomly horizontal mirroring during training. Both data augmentation and the selection of $\hat{\mathcal{P}_\triangle}$ take place at run-time, leading to different projected depth image for the same image across epoches, boosting the generalization ability of the network. And the training process aims to make the network regress to the selected transformation:

\begin{equation}
\label{equ:8}
    \mathcal{L}(\mathcal{P}_{\triangle,*}, \hat{\mathcal{P}_\triangle}) = \sum^{N_d}_{i=1} (\mathcal{L}_{t}(\mathcal{P}_{\triangle,i}, \hat{\mathcal{P}_\triangle}) + \mathcal{L}_{r}(\mathcal{P}_{\triangle,i}, \hat{\mathcal{P}_\triangle}) )
\end{equation}
where the cost function for translation $\mathcal{L}_{t}(\cdot, \cdot)$ is smooth L1 loss and the cost function for rotation is defined as the quaternion distance:

\begin{equation}
\begin{split}
    \label{equ:9}
    \mathcal{L}_{r}(\textbf{q}, \hat{\textbf{q}}) &= \Pi(\hat{\textbf{q}} \cdot \textbf{q}^{-1}) \\
    \Pi(\textbf{q}) =& \texttt{atan2}(\sqrt{qx^2+qy^2+qz^2}, |qw|)
\end{split}
\end{equation}
Different from original CMRNet \cite{cmrnet}, we add supervision on the prediction of each layer in the POET to enforce the stacked decoders to iteratively refine the \textit{pose queries}. Later experiments conclude that the estimated results are gradually optimized with the deepen of decoder layer.

\section{EXPERIMENTAL RESULTS}
\label{result}

\subsection{Setups}

We implemented the proposed work using PyTorch library. Aiming for a fair comparison, we integrated our work into CMRNet \cite{cmrnet} pipeline, so that the only difference in the experiments is the network itself. 
The proposed network is trained from scratch for 500 epoches using ADAM optimizer with default parameters, a batch size of 24 and an initial learning rate of $1e^{-4}$ on a single GeForce RTX 3090 GPU. 
We perform experiments on the KITTI odometry dataset \cite{kitti}. Sequences 03,05-09 are used for training while 00 is used for validation, and we also perform evaluation on KITTI 01,02,10,11,14,15 sequences so that the test map is never seen by the network during training.
We use LiDAR SLAM poses from \cite{cmrnet} as the ground-truth on training and validation sequences since the original KITTI poses cause map inconsistency in loop closures, and we directly use the original KITTI poses on evaluation sequences. 
During evaluation process,we add a transformation $\hat{\mathcal{P}_\triangle}$ on the ground-truth poses as initial poses $\mathcal{P}_0$ and thus the network is actually aim to estimate the randomly selected transformation.
We train three instances of the proposed network varying the select range of $\hat{\mathcal{P}_\triangle}$. For the network in the 1-st iteration, the range for the translation is $[-2m, +2m]$ and rotation is $[-10^\circ, +10^\circ]$. The sampling range for the 2-nd iteration and 3-rd iteration is $\pm1m/\pm2^\circ$ and $\pm0.6m/\pm2^\circ$, respectively. 
During depth image generation, we use a occlusion estimation filter \cite{occlusion} to discard occluded points. 
The proposed network contains $N_d=6$ decoder layers in the POET module in this work.

\subsection{Ablation Analysis}

\subsubsection{Multiple hypotheses aggregation}

We apply Mmultiple hypotheses aggregation method to \textit{pose queries} in the proposed DETR module to reduce the uncertainty of localization performance. To prove the effectiveness, we conduct the ablation studies that we test the proposed network with different number of \textit{pose queries}, that is, $N_q=1,5,10,15,20$, and evaluate the standard deviation of final localization errors over 10 runs. 
As shown in the Tab. \ref{table:N_q}, with the increase of $N_q$, the standard deviation of performance is reduced and the phenomenon can be observed in the network of both the 1-st and 3-rd iteration, which concludes the effectiveness of applied multiple \textit{pose queries} aggregation strategy. 
Note that this strategy does not need to re-train the network. 
It can be seen that the network with $N_q=20$ does not have significant improvement on localization performance compared to the one with $N_q=15$, so we set $N_q=15$ in later experiments for efficiency.

\begin{table}[t]
	\centering
	\caption{Performance disturbance of the proposed network with different number of pose queries.}
    \renewcommand\arraystretch{1.5}
	\begin{threeparttable}
		\begin{tabular}{c|c c| c c }
			\toprule
    			 & \multicolumn{2}{c|}{std. Mean error $\downarrow$} & \multicolumn{2}{c}{std. Median error $\downarrow$} \\
                $N_q$ & Trans. [cm] & Rot. [$^\circ$] & Trans. [cm] & Rot. [$^\circ$] \\
    			\midrule
                \multicolumn{5}{c}{iteration=1} \\
                \hline
       		1 & 0.3408 & 0.0109 & 0.1007 & 0.0016 \\
                5 & 0.2973 & 0.0094 & 0.0717 & 0.0015 \\
                10 & 0.2426 & 0.0082 & 0.0508 & 0.0014 \\
                15 & 0.2138 & 0.0060 & \textbf{0.0379} & 0.0010 \\
                20 & \textbf{0.1911} & \textbf{0.0058} & 0.0452 & \textbf{0.0008} \\
                \hline
                \multicolumn{5}{c}{iteration=3} \\
                \hline
                1 & 0.6126 & 0.0114 & 0.7250 & 0.0090 \\
                5 & 0.2512 & 0.0083 & 0.4259 & 0.0066 \\
                10 & 0.2341 & 0.0076 & 0.2787 & 0.0041 \\
                15 & \textbf{0.1178} & 0.0062 & \textbf{0.2028} & \textbf{0.0027} \\
                20 & 0.2281 & \textbf{0.0048} & 0.2822 & \textbf{0.0027} \\
			\bottomrule
		\end{tabular}
            \begin{tablenotes}
                \footnotesize
                \item[] The standard deviations (std.) are calculated over 10 runs.
            \end{tablenotes}
	\end{threeparttable}
	\label{table:N_q}
\end{table}

\subsubsection{Iterative optimization within a single network}

Each proposed POET module contains $N_d=6$ decoder layer in this work, we also provide the performance of the prediction from all the decoder layer in the 1-st iteration network to show the whole optimization process in each POET. As shown in the Tab. \ref{table:N_d}, the localization errors are constantly reduced with the deepen of decoder layer. It attribute to that each update take prior knowledge from the previous update, which make the update process more stable and fast. 

\begin{table}[t]
	\centering
	\caption{Localization performance of predicted poses from different layer in pose estimator Transformer module.}
    \renewcommand\arraystretch{1.5}
	\begin{threeparttable}
		\begin{tabular}{c|c c| c c }
			\toprule
			 & \multicolumn{2}{c|}{Mean error $\downarrow$} & \multicolumn{2}{c}{Median error $\downarrow$} \\
            depth & Trans. [cm] & Rot. [$^\circ$] & Trans. [cm] & Rot. [$^\circ$] \\
			\midrule
			0 & 182.0048 & 9.6583 & 187.0581 & 9.9386 \\
       		1 & 132.4100 & 5.7579 & 129.5515 & 5.6383 \\
                2 & 66.4576 & 1.7959 & 55.3583 & 1.5146 \\
                3 & 55.0531 & 1.7134 & 44.2252 & 1.4639 \\
                4 & 52.2464 & 1.6618 & 41.7836 & 1.4386 \\
                5 & 51.5909 & 1.6353 & 41.2128 & 1.4065 \\
                6 & \textbf{51.1117} & \textbf{1.6173} & \textbf{40.9964} & \textbf{1.3900} \\
			\bottomrule
		\end{tabular}
            \begin{tablenotes}
                \footnotesize
                \item[] The results are calculated over 10 runs.
            \end{tablenotes}
	\end{threeparttable}
	\label{table:N_d}
\end{table}

\subsubsection{Iterative optimization using multiple networks}
In this work, we also adopt the iterative optimization following CMRNet \cite{cmrnet} to obtain a better localization performance. In the Tab. \ref{table:iteration}, we show the localization performance of each optimization iteration using the proposed network. 
Firstly, we perform iterative optimization using the same network three times (`ours[1-st]' in the Tab. \ref{table:iteration}), it shows that the network can further get a better accuracy after twice optimization, but the results cannot be improved when optimize another time. The reason behind this should be that the generation of new projected depth image can provide a more ideal data to the network so that the network can estimate more accurately to some degree, but this strategy has a upper bound if the data distribution in evaluation scene was very different from training scene, \textit{e.g.}, $\pm200$cm translation and $\pm10^\circ$ rotation in training phase \textit{v.s.} $\pm60$cm translation and $\pm2^\circ$ rotation in evaluation phase. The trained network cannot ideally adopt to a scene with such different data distribution. 
Therefore, we also test the iterative optimization using three networks trained under different select range of $\hat{\mathcal{P}_\triangle}$ as mentioned before (`ours[full]' in the Tab. \ref{table:iteration}). Clearly, the performance of the proposed network is further optimized. It concludes multiple network can further boost the upper-bound performance of the localization method that a single network is hard to achieve.

\begin{table*}[!h]
    \centering
    \caption{Image-to-LiDAR map localization performance on KITTI 00 sequence.}
    \renewcommand\arraystretch{1.5}
	\begin{threeparttable}
        \setlength{\tabcolsep}{1mm}{
		\begin{tabular}{c| c| c c c| c c c c }
			\toprule
			\multirow{2}{*}{iteration} & $\hat{\mathcal{P}_\triangle}$ range in training & \multicolumn{3}{c|}{Mean error (Trans. [cm]/Rot. [${}^\circ$]) $\downarrow$} & \multicolumn{4}{c}{Median error (Trans. [cm]/Rot. [${}^\circ$]) $\downarrow$} \\
             & Trans. [cm]/ Rot. [${}^\circ$]& CMRNet \cite{cmrnet}${}^*$ & ours[1-st] & ours[full] & CMRNet  \cite{cmrnet}${}^{\bigstar}$ & CMRNet  \cite{cmrnet}${}^*$ & ours[1-st] & ours[full] \\
			\midrule
			0 & - / -         & 182.01/9.66 & 182.01/9.66 & 182.01/9.66 & - / - & 187.06/9.94 & 187.06/9.94 & 187.06/9.94 \\
       		1 & [-200,+200]/[-10,+10] & 61.91/1.97  & 51.16/\textbf{1.62} & \textbf{51.05/1.62} & 51.00/\textbf{1.39} & 52.02/1.68 & 41.01/\textbf{1.39} & \textbf{41.00/1.39} \\
                2 & [-100,+100]/[-2,+2]   & 36.25/1.41  & 44.08/1.48 & \textbf{27.68/1.06} & 31.00/1.09 & 27.80/1.20 & 34.63/1.25 & \textbf{20.27/0.90} \\
                3 & [-60,+60]/[-2,+2]   & 26.29/1.09  & 44.87/1.50 & \textbf{25.19/0.91} & 27.00/1.07 & \textbf{19.25}/0.91 & 34.74/1.25 & 19.67/\textbf{0.79} \\
			\bottomrule
		\end{tabular}
            \begin{tablenotes}
                \footnotesize
                \item[1] The best performance is highlighted by \textbf{BOLD}. All the results are averaged over 10 runs.
                \item[$^{*}$] The results are obtained by open-sourced weights in \url{https://github.com/cattaneod/CMRNet}.
                \item[${}^{\bigstar}$] The results are directly obtained from its original publication so we do not know the initial localization errors.
            \end{tablenotes}
        }
	\end{threeparttable}
	\label{table:iteration}
\end{table*}

\subsection{Comparison with State-of-the-Arts}

Since the fair comparison must be conducted under the same initial pose error $\hat{\mathcal{P}_\triangle}$, we compare our proposed image-to-LiDAR map localization network to the only open-sourced CMRNet \cite{cmrnet} with the same settings. 
The comparative results are shown in the Tab. \ref{table:iteration} and Tab. \ref{table:compare}. Our proposed network achieves an excellent localization accuracy and outperforms CMRNet \cite{cmrnet} with a lot margin in most scene. It concludes that the proposed POET module is a much better pose estimator compared to the vanilla pose regressor \cite{cmrnet}.

Tab. \ref{table:compare} also shows that our proposal can achieve a significantly improved localization accuracy on varying scene of KITTI odometry dataset \cite{kitti}, starting from an initial rough pose $\mathcal{P}_0$ displaced up to 3.4 m and 17${}^\circ$, which can meet the requirements of high-level autonomous driving. The network does not work well on KITTI-01 sequence since such a highway scene has very few geometry can be matched, but the localization errors are still reduced with a lot margin.

Nevertheless, it is worth to note that the proposed approach does not take advantage of neither odometry procedure nor multi-frame analysis, and the captured camera and the map in the evaluation scene are totally unseen by the network during training.
It shows a potential advantage of the proposed network that the network learns to match cross-modal data and then estimate pose regardless of the camera model and the scene map, indicating an excellent generalization ability.

\begin{figure*}[t]
    \centering
    \includegraphics[width=0.97\linewidth]{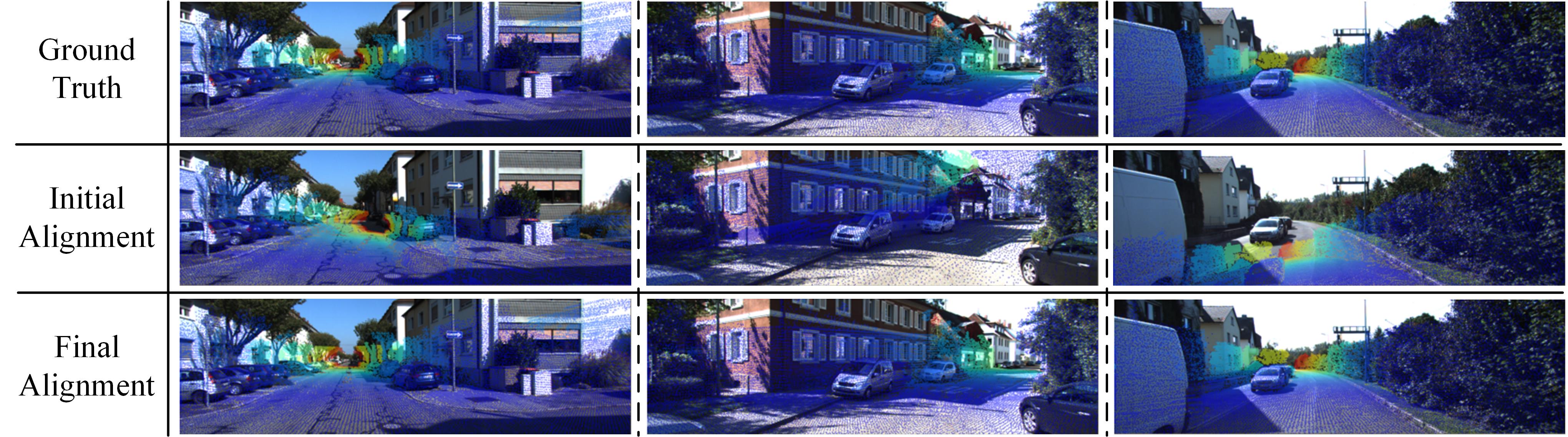}
    \caption{Visualization of some samples. The initial alignment is based on the initial pose $\mathcal{P}_0$, while the final alignment is using the result after three iteration with the proposed network. }
    \label{fig:align}
\end{figure*}

Fig. \ref{fig:align} visualizes some samples in the evaluation phase. Although the errors of alignments due to the initial pose $\mathcal{P}_0$ are large, the network can localize accurately and thus align the image to LiDAR map with a high precision.

\begin{table}[t]
    \centering
    \caption{Comparison with the baseline on various sequence of KITTI odometry dataset (KT).}
    \renewcommand\arraystretch{1.5}
	\begin{threeparttable}
        \setlength{\tabcolsep}{1mm}{
		\begin{tabular}{c| c c | c c }
			\toprule
			     \multirow{2}{*}{Dataset}  & \multicolumn{2}{c|}{CMRNet \cite{cmrnet}$^*$} & \multicolumn{2}{c}{ours [full]} \\
                               & Mean [cm/${}^\circ$] & Median [cm/${}^\circ$] & Mean [cm/${}^\circ$] & Median [cm/${}^\circ$] \\
			\midrule
                      KT00 & 26.29/1.09 & \textbf{19.25}/0.91 & \textbf{25.19/0.91} & 19.67/\textbf{0.79} \\
                      KT01 & 115.53/2.55 & 101.43/2.00 & \textbf{113.07/1.82} & \textbf{82.25/1.19} \\
                      KT02 & 63.50/1.63 & 43.01/1.29 & \textbf{60.10/1.45} & \textbf{34.86/1.13} \\
                      KT10 & 43.67/1.52 & 29.56/1.13 & \textbf{39.55/1.33} & \textbf{26.46/0.94}  \\
                      KT11 & 46.77/1.72 & 28.07/1.29 & \textbf{43.78/1.42} & \textbf{26.97/0.98} \\
                      KT15 & 25.86/0.91 & \textbf{18.21}/0.77 & \textbf{24.13/0.75} & 18.64/\textbf{0.75} \\
			\bottomrule
		\end{tabular}
            \begin{tablenotes}
                \footnotesize
                \item[1] The best performance is highlighted by \textbf{BOLD}. All the results are averaged over 10 runs.
                \item[$^{*}$] The results are obtained by open-sourced weights.
            \end{tablenotes}
        }
        \end{threeparttable}
\label{table:compare}
\end{table}

\subsection{Efficiency Evaluation}

Finally, we also test the size and running efficiency of the proposed network. The proposed POET only has few parameters compared to vanilla pose regressor in CMRNet \cite{cmrnet}, thus the overall model size is significantly reduced to about 1.1956 millions (M) versus 3.7116 M in CMRNet \cite{cmrnet}. Also, the proposed network can run about 67 frames per second (FPS) with $N_q=15$ \textit{pose queries}, which meet the real-time requirement in practical application.

\section{CONCLUSIONS}
\label{conclusion}

In this paper, we address the cross-modal localization by proposing a novel image-to-LiDAR map localization network. 
The network extracts image features and LiDAR features respectively and then calculate the cost volume between them as the image-to-map matching information. Then, the pose is implicitly represented as high-dimensional features, \textit{i.e.}, \textit{pose query} and updated by a proposed pose estimator module called POET. The update process is applied by constantly retrieving relevant information from the cost volume by attention mechanism in a Transformer architecture, while former update could provide prior knowledge to later update process so as to make the optimization more stable and fast. Moreover, to reduce the uncertainty caused by randomly initialized \textit{pose query}, we apply multiple hypotheses aggregation strategy in each POET to decrease the deviation of localization performance. 
The proposed localization network is fully analyzed on large-scale outdoor scene and concluded to be able to localize a monocular camera with a improved accuracy. The experiments proved the method could learn to match cross-modal data and estimate pose instead of learning the map, which is suitable for adopting to practical usage in high-level autonomous driving in varying scenarios.



\bibliographystyle{ieeetr}
\bibliography{ref}

\end{document}